\documentclass[journal]{IEEEtran}
\usepackage{booktabs}
\usepackage{multirow}
\usepackage{amsmath,amsfonts}
\usepackage[ruled,vlined, linesnumbered]{algorithm2e}
\usepackage{amsmath, amssymb}
\usepackage{algpseudocode}
\usepackage{array}
\usepackage[caption=false,font=normalsize,labelfont=sf,textfont=sf]{subfig}
\usepackage{textcomp}
\usepackage{stfloats}
\usepackage{url}
\usepackage{verbatim}
\usepackage{graphicx}
\usepackage{cite}  
\usepackage{multicol}
\usepackage{makecell}
\usepackage{lineno}
\usepackage{xcolor}
\usepackage{multirow}
\usepackage{arydshln}
\usepackage{threeparttable}
\usepackage{pifont} 
\usepackage{amssymb} 

\begin{document}

\bstctlcite{IEEEexample:BSTcontrol}

\title{Adapt as You Say: Online Interactive Bimanual Skill Adaptation via Human Language Feedback}

\author{Zhuo Li, Dianxi Li, Tao Teng, Quentin Rouxel, Zhipeng Dong, Dennis Hong,~\IEEEmembership{Member,~IEEE}, \\Darwin Caldwell,~\IEEEmembership{Fellow,~IEEE} and Fei Chen,~\IEEEmembership{Senior Member,~IEEE}
\thanks{*This work is supported in part by the Research Grants Council of the
Government of the Hong Kong SAR via the Grant 24209021, 14213324, C7100-22GF and in part by the InnoHK of the Government of the Hong Kong SAR via the Hong Kong Centre for Logistics Robotics. \textit{(Corresponding author: Fei Chen.)}}
\thanks{Zhuo Li, Dianxi Li, Tao Teng, Quentin Rouxel, Zhipeng Dong and Fei Chen are with the Collaborative and Versatile Robots (CLOVER) Laboratory, T-Stone Robotics Institute, The Chinese University of Hong Kong, Hong Kong (e-mail: zli@mae.cuhk.edu.hk; dianxili@cuhk.edu.hk; tao.teng@cuhk.edu.hk; quentinrouxel@cuhk.edu.hk; zhipengdong@cuhk.edu.hk; f.chen@ieee.org).}
\thanks{Dennis Hong is with the Department of Mechanical and Aerospace Engineering, University of California, Los Angeles, USA (e-mail: dennishong@g.ucla.edu).}
\thanks{Darwin Caldwell is with the Department of Advanced Robotics, Istituto Italiano di Tecnologia, Genoa, Italy (e-mail: darwin.caldwell@iit.it).}
}
\maketitle

\begin{abstract}
Developing general-purpose robots capable of autonomously operating in human living environments requires the ability to adapt to continuously evolving task conditions. However, adapting high-dimensional coordinated bimanual skills to novel task variations at deployment remains a fundamental challenge. In this work, we present BiSAIL (Bimanual Skill Adaptation via Interactive Language), a novel framework that enables zero-shot online adaptation of offline-learned bimanual skills through interactive language feedback. The key idea of BiSAIL is to adopt a hierarchical reason-then-modulate paradigm, which first infers generalized adaptation objectives from multimodal task variations, and then adapts bimanual motions via diffusion modulation to achieve the inferred objectives. Extensive real-robot experiments across six bimanual tasks and two dual-arm platforms demonstrate that BiSAIL significantly outperforms existing methods in human-in-the-loop adaptability, task generalization and cross-embodiment scalability. This work enables the development of adaptive bimanual assistants that can be flexibly customized by non-expert users via intuitive verbal corrections. Experimental videos and code are available at https://rip4kobe.github.io/BiSAIL/.
\end{abstract}


\section{Introduction}
\begin{figure}[!t]
\centering
\includegraphics[width=0.95\linewidth]{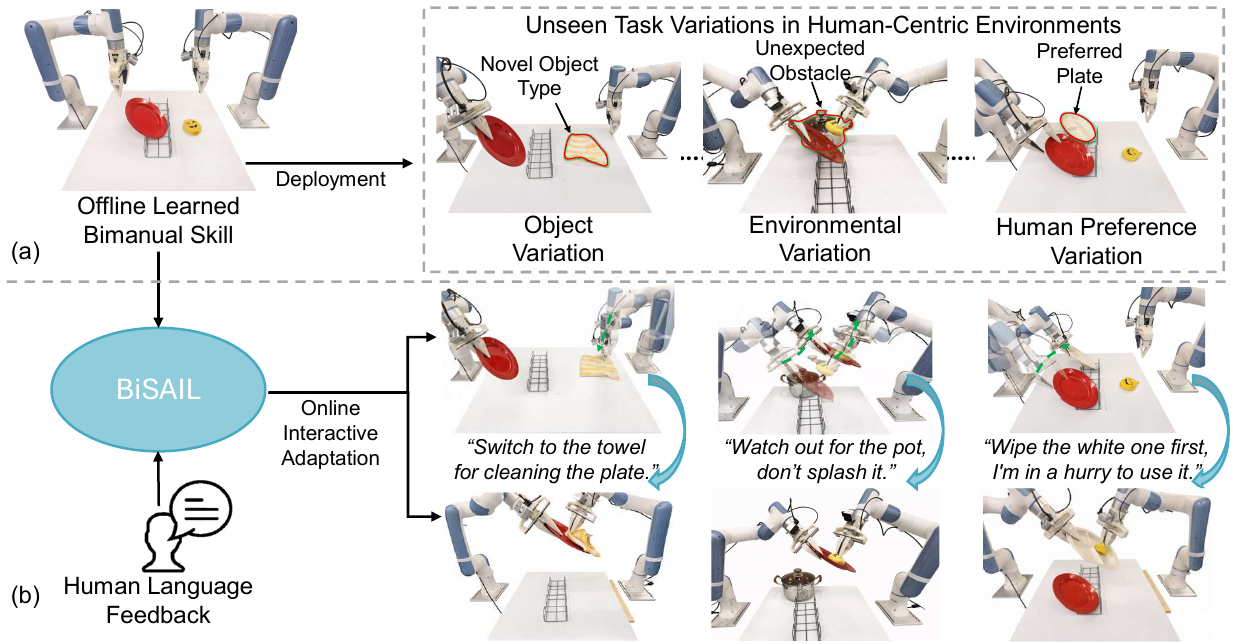}
\caption{\textbf{Illustration of Online Interactive Bimanual Skill Adaptation.} (a) Offline-learned bimanual skills often encounter diverse task variations when deployed in human-centric environments. (b) BiSAIL enables online interactive adaptation of learned bimanual skills through human language feedback, facilitating zero-shot generalization to unseen task variations.}
\label{fig1}
\end{figure}

\IEEEPARstart{A} hallmark of human manipulation intelligence is bimanual adaptability \cite{cui2021toward}. Humans not only excel at learning complex bimanual skills, but also continuously adapt to novel situations, respond effectively to diverse feedback, and dynamically coordinate dual-arm behaviors. Enabling comparable adaptability in robots is crucial for their practical deployment in unstructured human-centric environments, where task conditions are uncertain and ever-changing \cite{kemp2007challenges}. For instance, consider a personal robot tasked with cleaning dishes after a meal (see Figure \ref{fig1}). Instead of rigidly replicating an offline-learned wiping behavior, the robot must handle diverse task variations when operating alongside humans, such as adapting to alternative cleaning tools, avoiding unforeseen obstacles, or selecting dishes based on user preferences. This paper aims to advance robot bimanual adaptability by enabling online interactive adaptation of learned bimanual skills through human language feedback, a step toward building general-purpose robots that can be flexibly customized by non-expert users via intuitive verbal corrections. However, achieving such language-guided bimanual adaptability poses several challenges: (a) Interpreting implicit adaptation objective from ambiguous natural language feedback; (b) Modulating high-dimensional bimanual motions to accomplish the adaptation objective on-the-fly, while preserving dual-arm coordination and task compatibility; (c) Ensuring the accuracy and robustness of the adaptation process across diverse task variations.

Researchers have explored various skill adaptation strategies. A widely adopted approach is trajectory modulation \cite{gams2014coupling,li2025modified,franzese2023interactive,wang2024cooperative}, where manipulation skills are learned using Imitation Learning (IL)-based parametric models such as DMPs \cite{wang2024cooperative} or GPs \cite{franzese2023interactive}, and adapted to task variations by modulating generated trajectories.
However, these methods depend on explicitly defined adaptation objectives (e.g., via-points \cite{li2025modified}), making them ineffective in handling implicit intent conveyed through language feedback. 
Moreover, the limited expressiveness of parametric IL models constrains their ability to modulate high-dimensional bimanual motions. This often necessitates modeling each arm separately \cite{wang2024cooperative} and incorporating hand-crafted coupling terms \cite{gams2014coupling} to enforce coordination, which restricts flexibility in complex bimanual tasks involving diverse coordination patterns. An alternative approach addresses skill adaptation via policy fine-tuning \cite{guo2025srsa,yu2020learning,yokoyama2023asc,liu2025human,wagenmaker2025steering}, leveraging either reinforcement learning \cite{yokoyama2023asc,wagenmaker2025steering} or meta-learning \cite{yu2020learning} to update the parameters of a learned IL policy for handling task variations. However, these methods typically performs under an episodic adaptation paradigm, assuming the robot resets and retries after each episode~\cite{guo2025srsa}, which entails extensive trial-and-error interactions. This limits their applicability in human-centric scenarios, where robots must adapt on-the-fly to respond user feedback without long fine-tuning phases (see Figure \ref{fig1}). More recent approaches have explored end-to-end learning of skill adaptation from language feedback, training policies on large-scale language-annotated datasets to directly map verbal corrections to motion modifications~\cite{bucker2022latte,shi2024yell,sharma2022correcting,cui2023no,li2025language,li2025towards}. However, due to the inherent ambiguity of language, constructing datasets that comprehensively capture the diversity of verbal corrections across different task variations remains infeasible. As a result, these methods require extensive data collection~\cite{shi2024yell} and lack generalization to unseen task scenarios~\cite{bucker2022latte}. Furthermore, this end-to-end black-box approach tightly couples feedback understanding with motion modifications, hindering the interpretability and scalability of the adaptation process. 

\begin{figure*}[!t]
\centering
\includegraphics[width=0.95\linewidth]{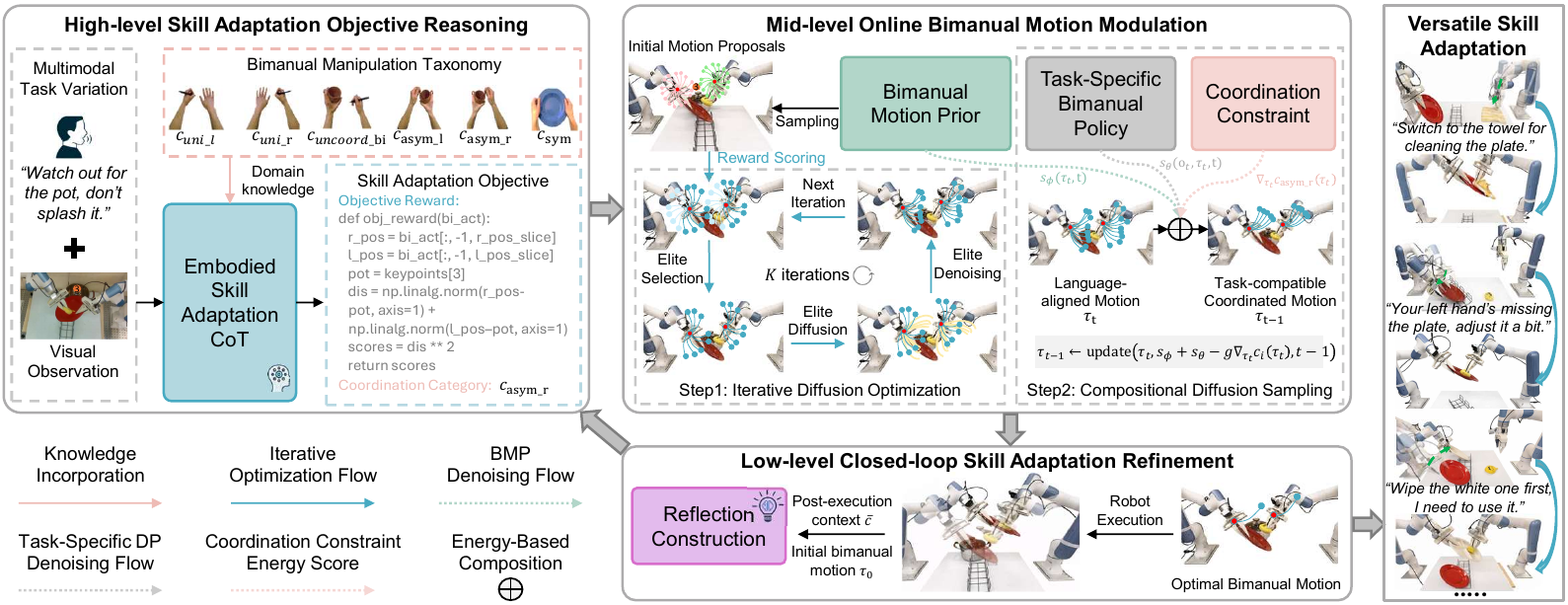}
\caption{\textbf{An Overview of BiSAIL.} (1) High-level Adaptation Objective Reasoning: ESA-CoT infers a structured bimanual adaptation objective from multimodal task variations; (2) Mid-level Bimanual Motion Modulation: Initial motion proposals sampled from BMP are first iteratively optimized toward the adaptation objective, and then modulated through compositional sampling to enforce dual-arm coordination
and task compatibility; (3) Low-level Skill Adaptation Refinement: A closed-loop reflection mechanism evaluates post-adaptation outcomes to refine both the adaptation objective and the resulting bimanual motion.}
\label{fig2}
\end{figure*}

As discussed above, existing methods remain insufficient to address the challenges of language-guided bimanual adaptability. We observe that human adaptation does not emerge from an end-to-end reflexive mechanism, but rather from a hierarchical paradigm that decouples adaptation objective reasoning (i.e., what to adapt) from bimanual motion modulation (i.e., how to adapt) \cite{kudithipudi2022biological}. Building on this insight, we introduce BiSAIL, an online interactive skill adaptation framework that enables zero-shot generalization of high-dimensional, coordinated bimanual skills to unseen task variations encountered during deployment. As shown in Figure \ref{fig2}, the core idea of BiSAIL is to emulate the hierarchical reason-then-modulate nature of human skill adaptation via three interconnected stages: \textbf{1) High-level Adaptation Objective Reasoning:} To interpret implicit adaptation intent from language feedback, we introduce an Embodied Skill Adaptation Chain-of-Thought (ESA-CoT) module that infers generalized bimanual adaptation objectives from multimodal task variations; \textbf{2) Mid-level Bimanual Motion Modulation:} To ground high-level adaptation objectives into executable bimanual motions, we propose a diffusion-based online motion modulation algorithm. It first aligns motion proposals from a learned Bimanual Motion Prior (BMP) with the adaptation objective via iterative diffusion optimization, followed by compositional diffusion sampling to enforce dual-arm coordination and task compatibility; \textbf{3) Low-level Skill Adaptation Refinement:} To improve accuracy and robustness, we incorporate a closed-loop refinement mechanism that augments the objective reasoning process with post-adaptation reflection, allowing refinement of both the adaptation objectives and the resulting bimanual motions.

Compared to existing end-to-end approaches, BiSAIL explores a novel hierarchical paradigm for bimanual skill adaptation that enables implicit intent interpretation and facilitates zero-shot generalization to unseen task variations. By decoupling feedback reasoning from motion modulation, BiSAIL improves adaptation interpretability, supporting error localization and correction via closed-loop iterative refinement. This decoupled design also allows plug-and-play integration with any offline-learned bimanual diffusion policy and diverse dual-arm embodiments, significantly enhancing framework scalability. Moreover, unlike prior methods that rely on trajectory modulation or policy fine-tuning, BiSAIL employs compositional motion sampling to flexibly integrate various coordination constraints and task requirements during online skill adaptation, eliminating the need for handcrafted coordination heuristics or task-specific model retraining. In summary, the contributions of this paper are as follows. 
\begin{itemize}
    \item {We introduce BiSAIL, an online interactive adaptation framework that enables zero-shot generalization of high-dimensional, coordinated bimanual skills to unseen task variations.}
    \item {We propose a diffusion-based online motion modulation algorithm that generates adapted bimanual motions which are simultaneously language-aligned, coordination-satisfied and task-compatible.}
    \item {We validate BiSAIL through extensive real-world experiments across diverse bimanual tasks and robot platforms, demonstrating strong language-guided adaptability, task generalization and cross-embodiment scalability.}
\end{itemize}

\section{Preliminary}
\subsection{Generative Diffusion Models}
Diffusion models are generative processes that learn a data distribution by reversing a fixed, gradual noising process. This process begins with the forward diffusion phase, where a data point $x_0$ from the true distribution $p(x)$ is progressively corrupted with Gaussian noise over $T$ timesteps. The noisy sample $x_t$ at any step $t$ can be expressed in a closed form as:
\begin{equation}
x_t = \sqrt{\bar{\alpha}_t}x_0 + \sqrt{1 - \bar{\alpha}_t}\epsilon, \quad \text{where } \epsilon \sim \mathcal{N}(0, I).
\end{equation}
To learn the reverse denoising process, a neural network $\epsilon_\theta$ is trained to predict the total added noise $\epsilon$ from the noisy input $x_t$ and timestep $t$. The training objective is to minimize the following loss function:
\begin{equation}
\label{eq2}
\mathcal{L}_{diff} = \mathbb{E}_{x_0 \sim p(x)} \left\| \epsilon_\theta \left( x_t, t \right) - \epsilon \right\|^2.
\end{equation}
In the denoising phase, starting with noisy pattern $x_T \sim \mathcal{N}(0, I)$, the trained model $\epsilon_\theta$ is used to iteratively denoise the sample from $x_T$ down to $x_0$:
\begin{equation}
x_{t-1} = \frac{1}{\sqrt{\alpha_t}} \left( x_t - \frac{1-\alpha_t}{\sqrt{1-\bar{\alpha}_t}} \epsilon_\theta(x_t, t) \right) + \sigma_t z.
\end{equation}

\subsection{Compositional Diffusion Sampling}
Diffusion models are closely connected to Energy-Based Models (EBMs) through score-based generative modeling \cite{du2023reduce}. The diffusion training objective in Equation \ref{eq2} is equivalent to learning a noise prediction network that approximates the score $s_\theta(x_t, t)$ of the perturbed data distribution $q(x_t)$:
\begin{equation}
\label{eq4}
s_\theta(x_t, t) = \nabla_{x_t} \log q(x_t) \approx -\frac{\epsilon_\theta(x_t, t)}{\sqrt{1 - \bar{\alpha}_t}}.
\end{equation}
This connection allows us to view the diffusion model as a noise-conditional EBMs, where an implicit energy function $E_\theta(x_t, t)$ is learned, satisfying:
\begin{equation}
    \nabla_{x_t} E_\theta(x_t, t) \propto -s_\theta(x_t, t) \propto \epsilon_\theta(x_t, t).
\end{equation}
This energy-based formulation facilitates compositional diffusion sampling, allowing the integration of scores from multiple independently learned distributions and external constraints to jointly guide data generation at test-time \cite{du2023reduce}, without requiring model retraining. Moreover, it supports the use of advanced score-based samplers, such as MCMC \cite{nijkamp2020anatomy}, to further enhance generation quality. A single step of MCMC sampler for compositional generation from two learned distributions $p(x_1)$ and $p(x_2)$ follows the update rule:
\begin{equation}
\label{eq5}
\begin{aligned}
x_t = x_t + \beta_t \left( s_{\theta_1}(x_t, t) + s_{\theta_2}(x_t, t) \right) + \sigma_t z. 
\end{aligned}
\end{equation}

\section{Problem Formulation}
\begin{figure}[!t]
\centering
\includegraphics[width=0.7\linewidth]{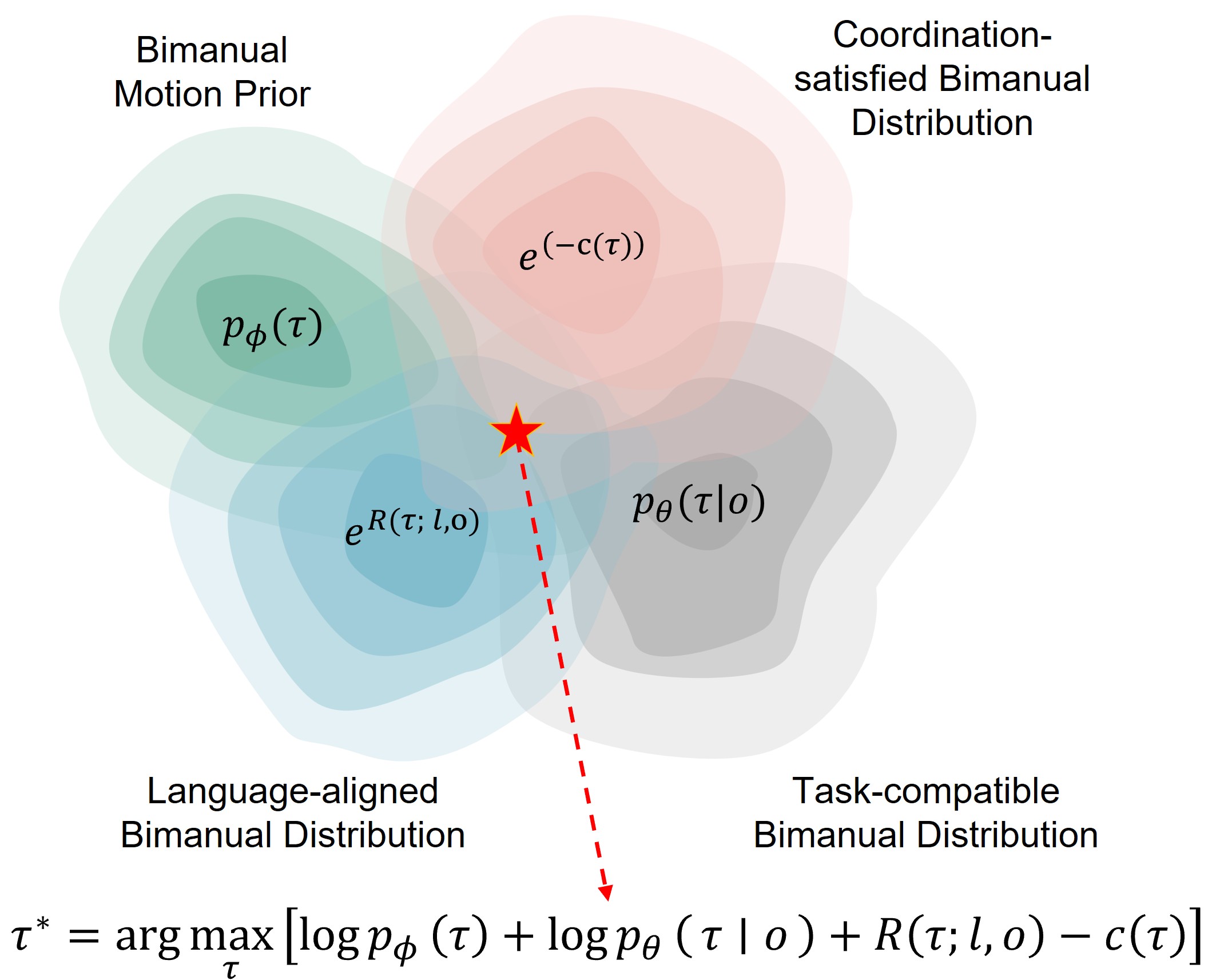}
\caption{\textbf{Problem Formulation.} We formulate the online interactive bimanual skill adaptation as a constrained probabilistic optimization problem. }
\label{fig3}
\end{figure}

We formulate the online interactive bimanual skill adaptation as a constrained probabilistic optimization problem. The objective is to obtain an optimal motion \( \tau^* \) from the bimanual motion prior \( p_\phi(\tau) \) that aligns with adaptation objective \( R(\tau; l, o) \), while satisfying offline-learned task-specific skill distribution \( p_\theta(\tau \mid o) \) and coordination constraint \( c(\tau) \). Formally, the problem is defined as:
\begin{equation}
\label{eq:optimization}
\begin{aligned}
\tau^* = \arg\max_{\tau \in \mathbb{R}^{N \times D}} \quad & R(\tau; l, o) \\
\text{s.t.} \quad & \tau \sim p_\phi(\tau) \\
& \tau \sim p_\theta(\tau \mid o) \\
& c(\tau) \leq \delta,
\end{aligned}
\end{equation}
where \(\tau = \{\tau_1, \tau_2, \dots, \tau_N\} \in \mathbb{R}^{N \times D} \) denotes a sequence of \( N \) dual-arm actions. Each action is a \( D \)-dimensional vector comprising the end-effector pose and gripper command for both arms. The objective reward \( R(\tau; l, o) \) evaluates how well a bimanual motion \( \tau \) satisfies the language feedback \( l \) under the given visual observation \( o \). The motion prior \( p_\phi(\tau) \) captures the distribution of all physically plausible bimanual motions within the dual-arm workspace. The offline-learned skill distribution \( p_\theta(\tau \mid o) \) encodes task-specific demonstrations and is instantiated using a diffusion policy \cite{chi2023diffusion,li2025manidp}. \( c(\tau) \) represents the coordination constraint that must be satisfied during adaptation. Directly solving Equation~\ref{eq:optimization} is intractable due to the implicit form of probabilistic constraints.
Therefore, we reformulate the problem as a Maximum A Posteriori (MAP) estimation, by defining a posterior distribution over bimanual motion $\tau$:
\begin{equation}
\begin{aligned}
p(\tau \mid l, o) \propto\, & p_\phi(\tau) \cdot p_\theta(\tau \mid o) \cdot \exp(R(\tau; l, o)) \\
& \cdot \exp(-c(\tau)).
\end{aligned}
\end{equation}
Maximizing this posterior is equivalent to maximizing its logarithm, which yields a more tractable formulation:
\begin{equation}
\label{eq8}
\begin{aligned}
\tau^* = \arg\max_{\tau \in \mathbb{R}^{N \times D}} \bigg[ & \log p_\phi(\tau) + \log p_\theta(\tau \mid o) \\
& + R(\tau; l, o) - c(\tau) \bigg].
\end{aligned}
\end{equation}
This formulation facilitates skill adaptation via compositional sampling (see Figure \ref{fig3}), where the optimal bimanual motion \( \tau^* \) is sampled from a composed energy-based distribution that integrates motion prior \( p_\phi(\tau) \) with adaptation objective \( R(\tau; l, o) \), task-specific policy \( p_\theta(\tau \mid o) \), and coordination constraint $c(\tau).$ 

\section{Method}
In this section, we first introduce adaptation objective reasoning and bimanual motion prior learning (Sections IV-A and IV-B). We then present the diffusion-based modulation algorithm, which first aligns the motion prior with the adaptation objective via iterative optimization, followed by compositional sampling for skill adaptation. Finally, we describe the closed-loop refinement mechanism that improves adaptation robustness.

\subsection{Skill Adaptation Objective Reasoning}\label{sec.III-B}
Human skill adaptation involves high-level reasoning over task variations \cite{kudithipudi2022biological}. Inspired by this, we propose ESA-CoT to infer bimanual adaptation objectives from multimodal task variations. Each infered objective includes: (i) a reward function $R(\tau; l, o)$ that encodes spatial constraints for adaptation, and (ii) a coordination category \(c_i\) that defines the required dual-arm interaction pattern during adaptation:
\begin{equation}
\{R(\tau; l, o),\; c_i \} = \mathcal{F}_{\text{ESA-COT}}\left(\Psi_{\text{MLLM}}(l, o\right)).
\end{equation}

\begin{figure}[!t]
\centering
\includegraphics[width=0.85\linewidth]{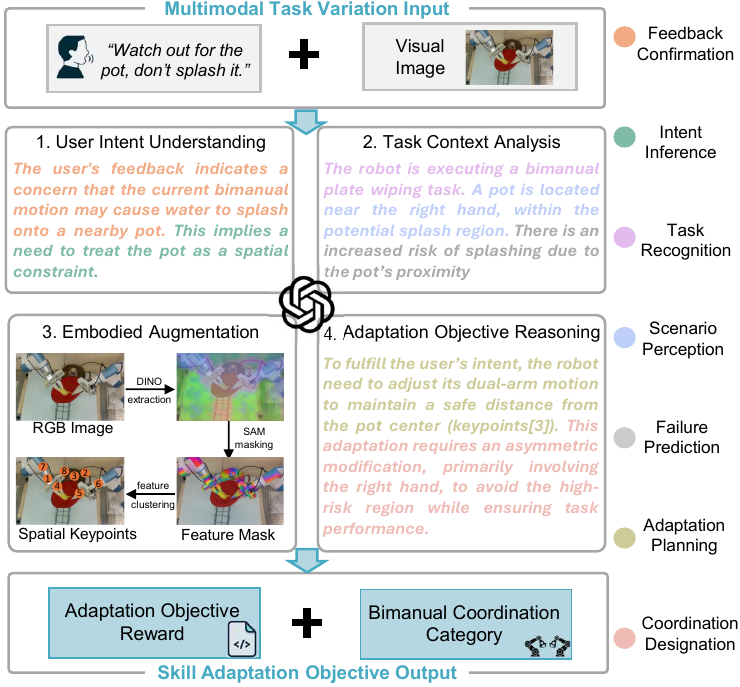}
\caption{\textbf{Bimanual Skill Adaptation Objective Reasoning.} ESA-CoT enables the objective reasoning through a structured chain-of-thought, augmented with robot embodied information and bimanual domain knowledge.}
\label{fig4}
\end{figure}

As illustrated in Figure~\ref{fig4}, ESA-CoT structures the objective reasoning process into four interleaved steps. It begins with \textit{User Intent Understanding}, where the Multimodal Large Language Model (MLLM) grounds the language feedback to align the inferred objective with the user’s underlying intent. Then, \textit{Task Context Analysis} is performed based on visual observation by identifying the task type, scenario relations, and potential failures. To enhance physical reliability, ESA-CoT incorporates an \textit{Embodied Augmentation} step \cite{zawalski2024robotic}, which integrates spatial keypoints of dual-arm end-effectors and target objects into the reasoning process via vision foundation models \cite{jose2025dinov2,ravi2024sam}. Finally, ESA-CoT performs \textit{Adaptation Objective Reasoning} to infer the objective reward \( R(\tau; l, o) \) and coordination category \( c_i \) based on the embodiment-augmented task context. Given the inherent ambiguity of language feedback, we represent \( R(\tau; l, o) \) as a non-differentiable black-box reward function \cite{yang2024diffusion}, which captures the goal-directed yet imprecise nature of human feedback while reducing the reasoning burden compared to a differentiable formulation (see Section~V-B). The inferred category $c_i$ is selected from the bimanual manipulation taxonomy~\cite{krebs2024formalization}, which defines three uncoordinated patterns (uncoordinated bimanual, unimanual left, and unimanual right) and three coordinated patterns (symmetric, right-dominant, and left-dominant):
\begin{equation}
\begin{aligned}
c_i \in \{&c_{uni\_l}, c_{uni\_r}, c_{uncoord\_bi}, \\
          &c_{asym\_l}, c_{asym\_r}, c_{sym}\} \subseteq \Phi_{taxonomy}.
\end{aligned}
\end{equation}

\subsection{Bimanual Motion Prior Learning}\label{sec.III-C}
Previous methods typically adapt skills within task-specific motion distributions~\cite{gams2014coupling}, limiting adaptation to a narrow vicinity of demonstrations. To overcome this, we learn a task-agnostic bimanual motion prior that models a broad distribution over physically plausible motions across the dual-arm workspace. The learned BMP enables bimanual adaptation beyond the original demonstrations
by supplying diverse and feasible motion proposals
for online modulation, thereby supporting more
versatile skill adaptation.

\begin{figure}[!t]
\centering
\includegraphics[width=0.9\linewidth]{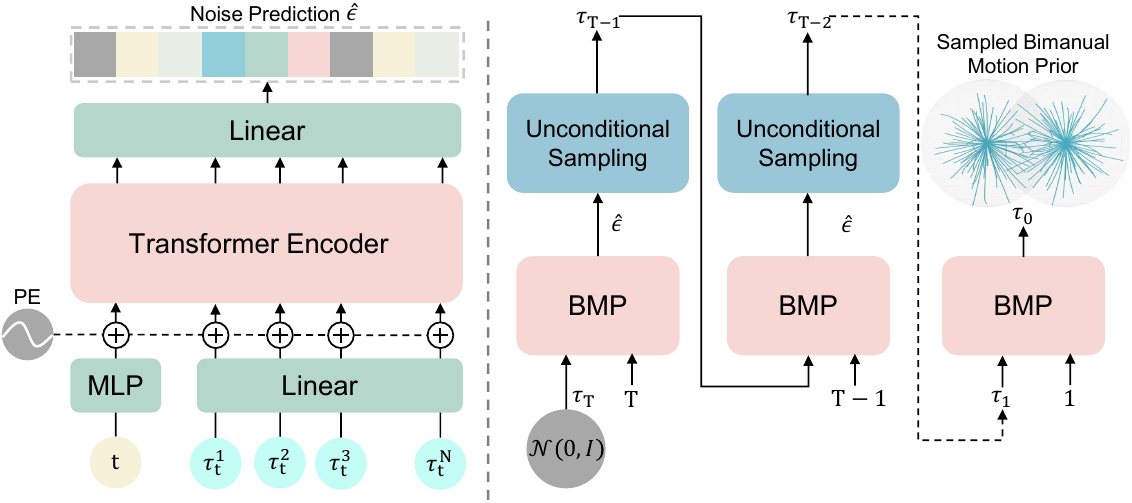}
\caption{\textbf{Bimanual Motion Prior Learning.} BMP is implemented as an unconditional diffusion model with Transformer encoder-only architecture.}
\label{fig5}
\end{figure}

\textbf{Model Architecture.} As shown in Figure~\ref{fig5}, BMP is implemented as an unconditional diffusion model \( \pi_\phi(\tau) \) with a Transformer encoder-only architecture~\cite{tevet2022human}. It takes as input a bimanual motion sequence $\tau$ and a diffusion timestep \( t \). The Transformer encoder then predicts the noise \( \hat{\epsilon} \) injected during the forward diffusion process for denoising. To ensure diversity and kinematic feasibility, we supplement the standard diffusion loss \( \mathcal{L}_{\text{diff}} \) with two auxiliary terms. First, \( \mathcal{L}_{\text{diversity}} \) encourages BMP to explore diverse motions by maximizing the average pairwise distance:
\begin{equation}
    \mathcal{L}_{\text{diversity}} = -\frac{1}{B(B - 1)} \sum_{i \ne j} \left\| \hat{\tau_{0}}^{(i)} - \hat{\tau_{0}}^{(j)} \right\|_2^2,
\end{equation}
where \( B \) denotes the batch size, and \( i \), \( j \) index different samples within the batch. $\hat{\tau_{0}}$ is the estimated clean bimanual motion derived from the noise prediction $\hat{\epsilon}$:
\begin{equation}
\label{eq13}
\hat{\tau}_0 = \frac{1}{\sqrt{\bar{\alpha}_t}} \left( \tau_t - \sqrt{1 - \bar{\alpha}_t} \, \hat{\epsilon} \right).
\end{equation}
Second, \( \mathcal{L}_{\text{smooth}} \) penalizes abrupt changes in motion and enforces second-order temporal continuity:
\begin{equation}
\mathcal{L}_{\text{smooth}} = \sum_{n=2}^{N-1} \left\| \hat{\tau_{0}}^n - 2\hat{\tau_{0}}^{n-1} + \hat{\tau_{0}}^{n-2} \right\|_2^2  .
\end{equation}
The overall training objective is a weighted combination of all components:
\begin{equation}
\mathcal{L}_{\text{total}} = \mathcal{L}_{\text{diff}} + \lambda_{1} \mathcal{L}_{\text{diversity}} + \lambda_{2} \mathcal{L}_{\text{smooth}},
\end{equation}
where \( \lambda_{1} \) and \( \lambda_{2} \) are hyperparameters controlling the relative importance of each auxiliary term. At sampling, a noisy bimanual motion \( \tau_T \sim \mathcal{N}(0, I) \) is sampled and iteratively denoised from \( T \) down to \( 1 \), using the predicted noise \( \hat{\epsilon} \) at each step to generate the final motion sequence \( \tau_0 \).

\begin{figure}[!t]
\centering
\includegraphics[width=0.9\linewidth]{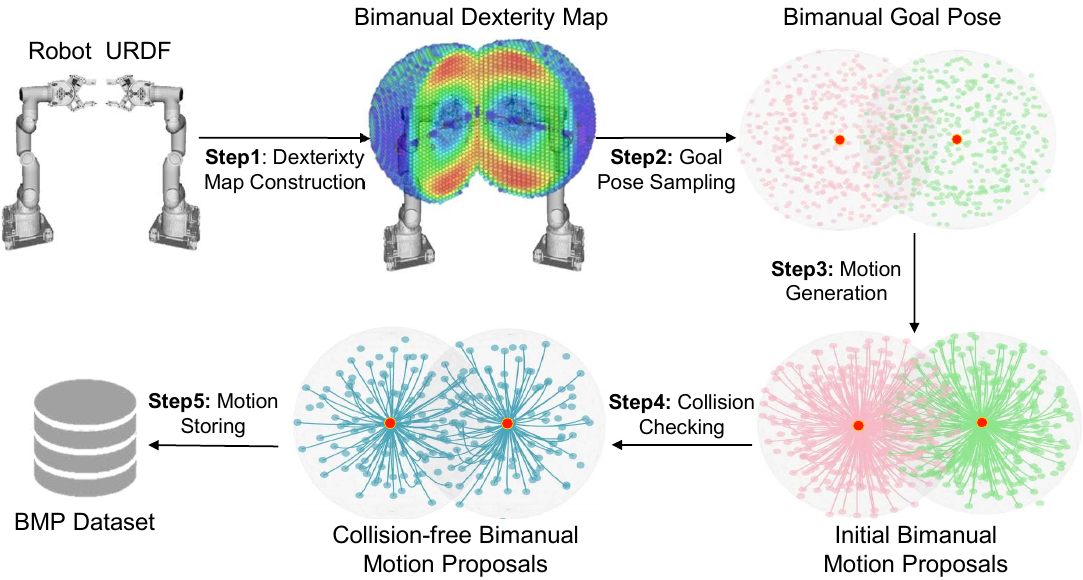}
\caption{\textbf{BMP Dataset Generation.} Self-supervised bimanual motion data generation pipeline for training BMP.}
\label{fig6}
\end{figure}

\textbf{Data Generation.} 
We develop a self-supervised data generation pipeline that constructs a large-scale dataset of diverse, valid, and collision-free bimanual motions for training BMP. As illustrated in Figure~\ref{fig6}, the pipeline begins by parsing the robot URDF to construct a bimanual dexterity map~\cite{quan2022dexterity}, which discretizes the dual-arm workspace into a voxel grid and evaluates a set of end-effector poses around each voxel center using the dexterity metric \( \omega \):
\begin{equation}
    \omega = \sqrt{\det\left(J J^{\top}\right)},
\end{equation}
where \( J \) denotes the Jacobian matrix of the corresponding arm. Based on the constructed dexterity map, initial dual-arm poses are selected as those with the highest dexterity scores across the workspace. We then uniformly sample goal poses around the initial dual-arm poses using Fibonacci sampling, incorporating orientation perturbations to enhance spatial diversity. The kinematic reachability of each sampled goal pose is verified by querying the precomputed dexterity map. For each valid initial-goal pose pair, we generate smooth bimanual motions via joint-space quintic interpolation. These generated motions are then filtered through collision checking to eliminate motions with inter-arm and self-collisions. Finally, all feasible bimanual motions are stored in the dataset using a relative format, represented as incremental pose displacements from the start to the goal pose. In total, we generate a dataset comprising 1000,000 bimanual motions for training BMP.
Since the data generation pipeline is fully parameterized by the robot's URDF, it can be seamlessly adapted to different dual-arm embodiments by simply replacing the input URDF. This enables scalable motion prior synthesis and facilitates cross-embodiment generalization, as demonstrated in Section V-F.

\subsection{Online Bimanual Motion Modulation}\label{sec.III-D}
We propose a diffusion-based online modulation algorithm to generate bimanual motions that are (i) language-aligned, consistent with the adaptation objective $R(\tau; l, o)$; (ii) task-compatible, feasible within the offline-learned task-specific skill distribution $p_\theta(\tau \mid o)$; and (iii) coordination-satisfied, adhering to the inferred coordination pattern $c_i$. The modulation proceeds in two stages: it first aligns motion proposals from BMP with the adaptation objective via iterative diffusion optimization, then performs compositional diffusion sampling to incorporate task requirements and coordination constraints.

\textbf{Iterative Diffusion Optimization}. 
To handle the non-differentiable nature of $R(\tau; l, o)$, we design a gradient-free iterative diffusion optimization strategy inspired by \cite{zhang2024diffusion}. As shown in Figure \ref{fig2}, it begins by sampling $M$ diverse bimanual motions from BMP to initialize the proposal set $A_0$:
\begin{equation}
    A^0 = \{\tau^i \sim p_\phi(\tau) \mid i = 1, \ldots, M\}.
\end{equation}
We then performs an iterative search loop to evaluate and mutate the initial proposal set using $R(\tau; l, o)$. Specifically, at each iteration $k$, we score the proposal set $\{R(\tau^i; l, o) \mid \tau^i\in A^{k-1} \}_{i=1}^M$ and select high-scoring elite proposals $E^{k}\subseteq A^{k-1}$:
\begin{equation}
q(\tau) = \frac{\exp\left( \zeta R(\tau; l, o) \right)}
             {\sum_{i=1}^{M} \exp\left( \zeta R(\tau^i; l, o) \right)}
\end{equation}

\begin{equation}
E^{k} = \left\{\tau^i {\sim}  q(\tau) \right\}_{i=1}^{M},
\end{equation}
where $\zeta$ is a tunable temperature parameter controlling the sharpness of $q$. To enhance exploration during search, we introduce mutations to the elite set $E^{k}$ via a truncated diffusion-denoising process. Specifically, we run the first $\mu$ steps of the forward diffusion process to obtain noised elite proposals $\bar{E}^{k}$:
\begin{equation}
\bar{E}^{k} = \left\{ \sqrt{\bar{\alpha}_\mu} \tau + \sqrt{1 - \bar{\alpha}_\mu} \, \epsilon \,\middle|\, \tau \in E^{k} \right\}.
\end{equation}
We then perform the final \( \mu \) steps of the denoising process to obtain the refined proposal set \( A^{k} \) for propagation.
\begin{equation}
\label{eq21}
A^{k} = \left\{ \tau \sim p_\phi(\tau \mid \bar{\tau}) \,\middle|\, \bar{\tau} \in \bar{E}^{k} \right\}.
\end{equation}

\begin{figure}[!t]
\centering
\includegraphics[width=0.95\linewidth]{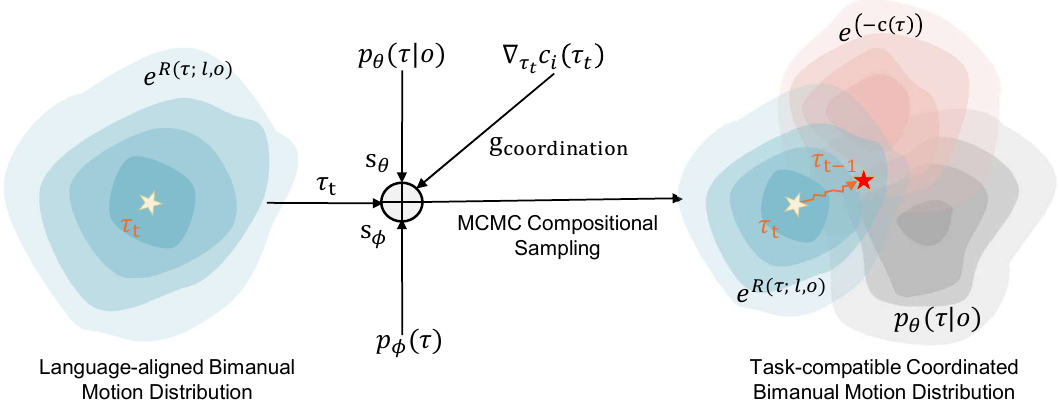}
\caption{\textbf{Bimanual Motion Modulation via Compositional Sampling.}
We employ MCMC sampling to optimize the joint motion distribution over adaptation objective, task-specific policy and coordination constraint.}
\label{fig7}
\end{figure}

\textbf{Compositional Diffusion Sampling.} 
Beyond aligning with the language-guided adaptation objective, the modulated bimanual motion must also satisfy task compatibility and coordination constraints. As shown in Figure~\ref{fig7}, we perform compositional sampling into each truncated denoising step to further optimize the composition of task distribution \( p_\theta(\tau \vert o) \) and coordination constraint \( c_i(\tau) \):
\begin{equation}
\label{eq22}
A^{k} = \left\{ \tau \sim p_\phi(\tau \mid \bar{\tau}) \cdot p_\theta(\tau \mid o) \cdot \exp(-c_i(\tau)) \,\middle|\, \bar{\tau} \in \bar{E}^{k} \right\}.
\end{equation}
Specifically, the noise prediction of \( \varepsilon_{\theta}(o, \tau_t, t) \) and \( \varepsilon_{\phi}(\tau_t, t) \) can be transformed into score estimation via Equation \ref{eq4}:
\begin{equation}
\begin{split}
s_{\theta} &= \nabla_{\tau_t} \log p(\tau_t \vert o) \approx -\frac{\varepsilon_{\theta}(o, \tau_t, t)}{\sqrt{1 - \bar{\alpha}_t}} \\
s_{\phi} &= \nabla_{\tau_t} \log p(\tau_t) \approx -\frac{\varepsilon_{\phi}(\tau_t, t)}{\sqrt{1 - \bar{\alpha}_t}}.
\end{split}
\end{equation}
The coordination constraint \( c_i(\tau_t) \) can also be modeled as an energy function that defines an EBM distribution $q(\tau_t)$:
\begin{equation}
q(\tau_t) \propto \exp\left(-c_i(\tau_t)\right).
\end{equation}
Notably, $c_i(\tau_t)$ is applied only when the inferred category $c_i$ is one of the three coordinated patterns (see Section V-A), which are designed as follows:
\begin{equation}
\begin{aligned}
c_{\text{sym}}(\tau_t) &= \sum_{n=1}^{N} \left\| \text{Log} \left( (\hat{\tau}_0^L)_n^{-1} \cdot (\hat{\tau}_0^R)_n \cdot (\tau_{\text{rel}})^{-1} \right) \right\|^2, \\
c_{\text{asym\_l}}(\tau_t) &= \sum_{n=1}^{N} \left\| \text{Log} \left( (\hat{\tau}_0^L)_n^{-1} \cdot (\hat{\tau}_0^R)_n \cdot \tau_{\text{rel}} \right) \right\|^2, \\
c_{\text{asym\_r}}(\tau_t) &= \sum_{n=1}^{N} \left\| \text{Log} \left( (\hat{\tau}_0^R)_n^{-1} \cdot (\hat{\tau}_0^L)_n \cdot \tau_{\text{rel}}\right) \right\|^2,
\end{aligned}
\end{equation}
where $\hat{\tau}_0^L$ and $\hat{\tau}_0^R$ denote the left and right end-effector poses of the estimated clean bimanual motion $\hat{\tau}_0$ at diffusion step $t$ (see Equation~\ref{eq13}). $\tau_{\text{rel}}$ is the reference relative pose computed from the original dual-arm configuration before adaptation. Finally, we sample modulated motions that satisfy all objectives via annealed MCMC \cite{du2023reduce} at each denoising step:
\begin{equation}
\label{eq:mcmc_update}
\tau_t \leftarrow \tau_t + \gamma_t\left(s_{\phi} + s_{\theta} - g \nabla_{\tau_t}c_i(\tau_t)\right) + \sigma_t z,
\end{equation}
where \( \gamma_t \) is annealing coefficient and $g$ is guiding ratio. This enables the bimanual modulation that are simultaneously language-aligned, task-compatible, and coordination-satisfied. The full procedure is summarized in Algorithm \ref{alg:alg1}.

\begin{algorithm}[t]
\caption{Online Bimanual Motion Modulation}
\small
\label{alg:alg1}
\KwIn{BMP model $\varepsilon_{\phi}(\tau_t, t)$, DP model $\varepsilon_{\theta}(o, \tau_t, t)$, adaptation objective reward $R(\tau; l, o)$, coordination constraint $c_i$.}
\KwOut{Adapted bimanual motion $\tau^\star$}
Initial proposals: $A^0 = \{\tau^i\}_{i=1}^M \sim p_\phi(\tau)$ \;
\For{$k = 1$ \KwTo $K$}{
    Proposal scoring: $\{R(\tau^i; l, o)|\tau^i\in A^{k-1} \}_{i=1}^M$
    
    Compute score distribution: 
    \begin{equation*}
    q(\tau) = \frac{\exp\left( \zeta R(\tau; l, o) \right)}
             {\sum_{i=1}^{M} \exp\left( \zeta R(\tau^i; l, o) \right)}
    \end{equation*}
    
    Elite selection: $E^{k} = \left\{\tau^i {\sim} q(\tau) \right\}_{i=1}^{M}$
    
    Elite diffusion: $\bar{E}^{k} = \left\{ \sqrt{\bar{\alpha}_\mu} \tau + \sqrt{1 - \bar{\alpha}_\mu} \, \epsilon \,\middle|\, \tau \in E^{k} \right\}$ 
    
    Elite denoising:
    
    \For{MCMC sampling steps $j \in \{1, \ldots, J\}$}{
        $\displaystyle s_{\phi} = \nabla_{\tau_t} \log p_\phi(\tau_t) \approx -\frac{\varepsilon_{\phi}(\tau_t, t)}{\sqrt{1 - \bar{\alpha}_t}}$ \;
        $\displaystyle s_{\theta} = \nabla_{\tau_t} \log p_\theta(\tau_t \vert o) \approx -\frac{\varepsilon_{\theta}(o, \tau_t, t)}{\sqrt{1 - \bar{\alpha}_t}}$ \;
        \eIf{$j < J$}{
            $\tau_t \leftarrow \text{update}(\tau_t, s_{\phi} + s_{\theta} - g \nabla_{\tau_t} c_i(\tau_t), t)$
        }{
            $\tau_{t-1} \leftarrow \text{update}(\tau_{t}, s_{\phi} + s_{\theta} - g \nabla_{\tau_t} c_i(\tau_t), t-1)$
        }
    } \
    Proposal propagation: $A^k = \{\tau_0^i \mid i = 1, \ldots, M\}$
}
$\tau^\star \gets \arg\max_{\tau \in A^K} R(\tau; l, o)$
\end{algorithm}

\subsection{Closed-loop Skill Adaptation Refinement}\label{sec.III-E}
We introduce a closed-loop refinement mechanism that corrects both the adaptation objective and the resulting bimanual motion. Specifically, we augment the original ESA-CoT reasoning with a reflection step~\cite{Ma2023EurekaHR}, where the MLLM is prompted with four key elements: the initial bimanual motion $\tau_0$, the modulated bimanual motion $\tau^\star$, post-execution task context $\bar{c}=(l, \bar{o})$, and the reasoning history $\mathcal{H}$ from the previous ESA-CoT step. Given this reflection-augmented input, the MLLM serves as a self-critic to refine the adaptation objective reward $R(\tau; l, o)$ and produce the adaptation success indicator $\delta \in \{0, 1\}$:
\begin{equation}
    \{R'(\tau; l, o),\; \delta\}
 = \mathcal{F}_{\text{ESA-CoT}}(\Psi_{\text{MLLM}}(\tau_0, \tau^\star, \bar{c}, \mathcal{H})).
\end{equation}
This closed-loop refinement continues until adaptation success (i.e., $\delta = 1$) or a maximum number of retries is reached, ensuring both the accuracy and robustness of skill adaptation.

\section{Experiments}
This section presents extensive experiments to evaluate the proposed BiSAIL framework. We aim to 1) analyze the effectiveness and design choices of each individual component within BiSAIL, 2) assess the overall performance of BiSAIL for online interactive bimanual skill adaptation under diverse real-robot tasks, 3) validate the scalability of BiSAIL across different robot embodiments.

\subsection{Setup}
\textbf{Implementation Details:}
The ESA-CoT module is implemented using GPT-4o with a temperature of 0.2. For BMP training, we use the AdamW optimizer with a learning rate of 1e-4 and a batch size of 256. During online motion modulation, the initial population size is set to 32. Both iterative search and MCMC sampling steps are set to 4, balancing exploration and efficiency. The number of denoising steps for BMP sampling and online modulation is set to 10. All experiments are conducted on a workstation equipped with an NVIDIA RTX 4090 GPU.

\begin{figure}[!t]
\centering
\includegraphics[width=0.85\linewidth]{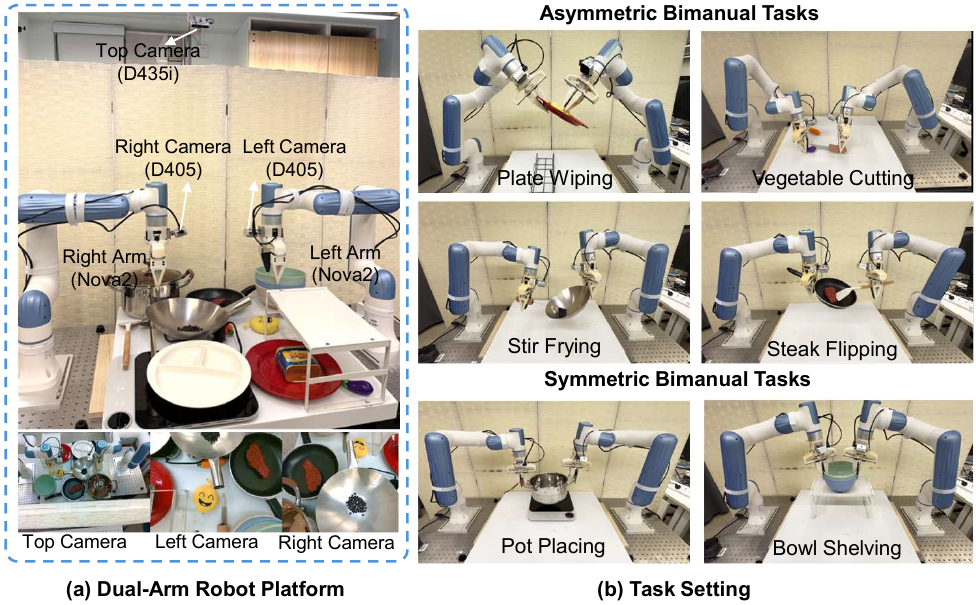}
\caption{\textbf{Experimental Setup.} (a) Overview of the dual-arm robot platform. (b) Evaluation tasks comprising six bimanual scenarios.}
\label{real_robot_experiments_setting}
\end{figure}

\textbf{Task Settings:}
The proposed framework is verified on the dual-arm robotic system DOBOT X-Trainer, which comprises two 6-DoF Nova2 robot arms equipped with 1-DoF grippers. Six representative  bimanual tasks are selected for the experiment (see Figure \ref{real_robot_experiments_setting}), including: 1) Four asymmetric tasks, where the two arms performed different functional roles in coordination; 2) Two symmetric tasks, which required synchronized coordination between both arms. For each task, the corresponding bimanual skill is learned  offline with Diffusion Policy (DP) \cite{chi2023diffusion}, which models the task-specific motion distribution $p_\theta(\tau \mid o)$. 

\begin{figure}[!t]
\centering
\includegraphics[width=0.9\linewidth]{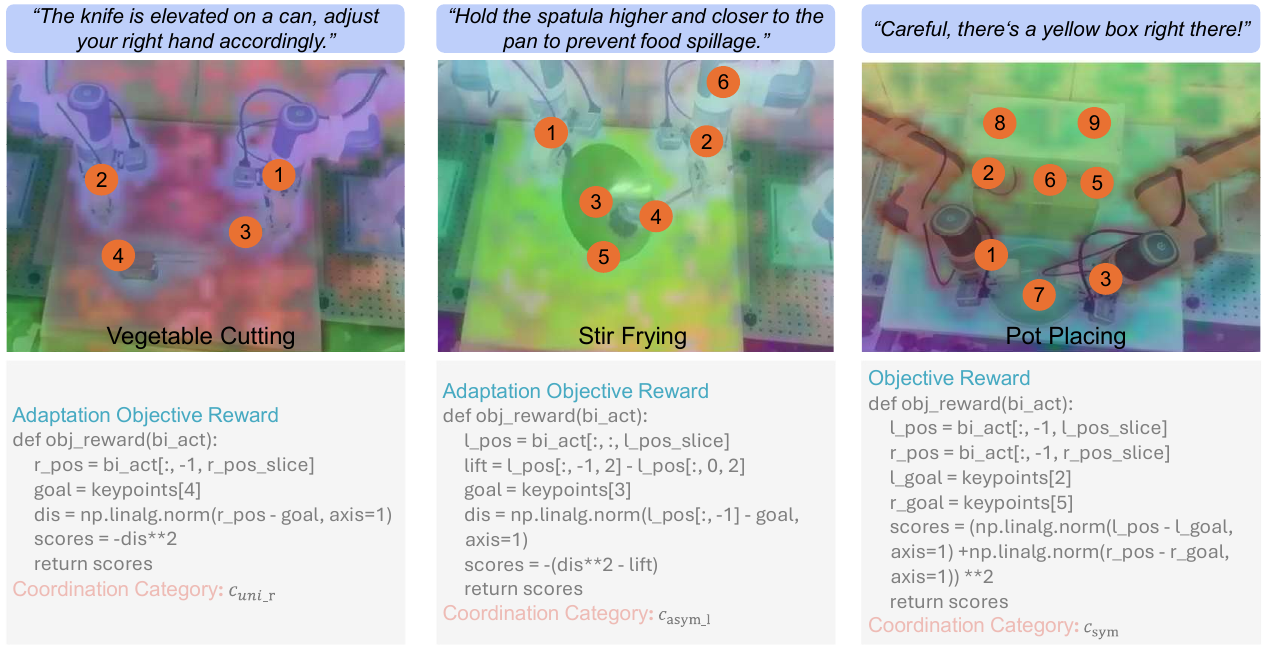}
\caption{\textbf{Qualitative Results of Adaptation Objective Reasoning.} ESA-CoT infers feasible rewards and coordination across diverse task variations.}
\label{ESA-COT qualitatives}
\end{figure}

\subsection{Evaluation of Adaptation Objective Reasoning}
To assess the designed ESA-CoT module, we construct a bimanual adaptation dataset containing 120 multimodal task variations across six selected tasks, where each variation combines language feedback with corresponding visual observations. Ground-truth adaptation objectives are annotated by human experts. We compare ESA-CoT against Few-shot Prompting~\cite{ma2023fairness} and Vanilla CoT~\cite{wei2022chain}. To assess the contribution of each reasoning step, five ablated variants are also evaluated (see Table~\ref{ESA-COT quantitatives}). Three metrics are used for quantitative evaluation: 1) Reasoning Accuracy (RA), which measures the correctness of the inferred adaptation objectives; 2) Executability Rate (ER), which quantifies the proportion of generated adaptation rewards that are programmatically executable; 3) ROUGE-L \cite{lin2003rouge}, which computes the linguistic similarity between the generated reasoning chains and human-annotated ground-truth. Each method is provided with 20 examples as demonstrations and evaluated on the remaining 100 examples in a zero-shot setting. 

\begin{table}[t]
\caption{Quantitative results of adaptation objective reasoning}
\label{ESA-COT quantitatives}
\centering
\scriptsize
\begin{threeparttable}
\begin{tabular*}{.5\textwidth}{@{\extracolsep{\fill}} lccc}
\toprule
Method & RA (\%) $\uparrow$ & ER (\%) $\uparrow$ & ROUGE-L $\uparrow$ \\
\midrule
Few-shot Prompting & 25.3 \scriptsize$\pm$ 1.2 & 17.6 \scriptsize$\pm$ 1.5 & N/A \\
Vanilla CoT & 44.5 \scriptsize$\pm$ 0.6 & 38.7 \scriptsize$\pm$ 0.4 & 0.55 \scriptsize$\pm$ 0.08 \\
ESA-COT (w/o UIU) & 85.7 \scriptsize$\pm$ 0.6 & 79.4 \scriptsize$\pm$ 0.7 & 0.88 \scriptsize$\pm$ 0.05 \\
ESA-COT (w/o TCA) & 80.1 \scriptsize$\pm$ 0.7 & 75.7 \scriptsize$\pm$ 0.2 & 0.77 \scriptsize$\pm$ 0.03 \\
ESA-COT (w/o EA) & 58.7 \scriptsize$\pm$ 0.5 & 49.6 \scriptsize$\pm$ 1.1 & 0.68 \scriptsize$\pm$ 0.06 \\
ESA-COT (w/o AOR) & 75.4 \scriptsize$\pm$ 0.7 & 69.7 \scriptsize$\pm$ 0.9 & 0.67 \scriptsize$\pm$ 0.07 \\
ESA-COT (w/o BMT) & 65.1 \scriptsize$\pm$ 1.1 & 62.3 \scriptsize$\pm$ 0.7 & 0.61 \scriptsize$\pm$ 0.02 \\
ESA-COT (w/ Diff Reward) & 66.5 \scriptsize$\pm$ 0.6 & 42.9 \scriptsize$\pm$ 1.1 & 0.91 \scriptsize$\pm$ 0.04 \\
\midrule
ESA-COT (Full) & \textbf{93.7 \scriptsize$\pm$ 0.3} & \textbf{89.8 \scriptsize$\pm$ 0.4} & \textbf{0.94 \scriptsize$\pm$ 0.03} \\
\bottomrule
\end{tabular*}
\begin{tablenotes}[flushleft]
\scriptsize
\item[] UIU: User Intent Understanding; TCA: Task Context Analysis; EA: Embodied Augmentation; AOR: Adaptation Objective Reasoning; BMT: Bimanual Manipulation Taxonomy
\end{tablenotes}
\end{threeparttable}
\end{table}

\begin{table}[t]
\centering
\scriptsize
\caption{Quantitative results of BMP evaluation}
\label{tab2}
\setlength{\tabcolsep}{2pt} 
\begin{tabular}{lcccc}
\toprule
Method & FID $\downarrow$ & Diversity $\uparrow$ & Smoothness $\downarrow$ & Time (s) \\
\midrule
U-Net & 20.61 \scriptsize$\pm$ 0.07 & 0.58 & 0.04 \scriptsize$\pm$ 0.01 & \textbf{0.10} \\
Transformer (Dec-only) & 13.54 \scriptsize$\pm$ 0.13 & 0.73 & 0.12 \scriptsize$\pm$ 0.03 & 0.27 \\
Transformer (Enc-Dec) & 3.21 \scriptsize$\pm$ 0.05 & \textbf{0.94} & 0.03 \scriptsize$\pm$ 0.04 & 0.51 \\
BMP (w/o auxiliary) & 3.73 \scriptsize$\pm$ 0.02 & 0.91 & 0.09 \scriptsize$\pm$ 0.03 & 0.16 \\
\midrule
BMP (Ours) & \textbf{2.48 \scriptsize$\pm$ 0.02} & 0.93& \textbf{0.02 \scriptsize$\pm$ 0.06} & 0.16 \\
\bottomrule
\end{tabular}
\end{table}

\begin{figure*}[!t]
\centering
\includegraphics[width=0.9\linewidth]{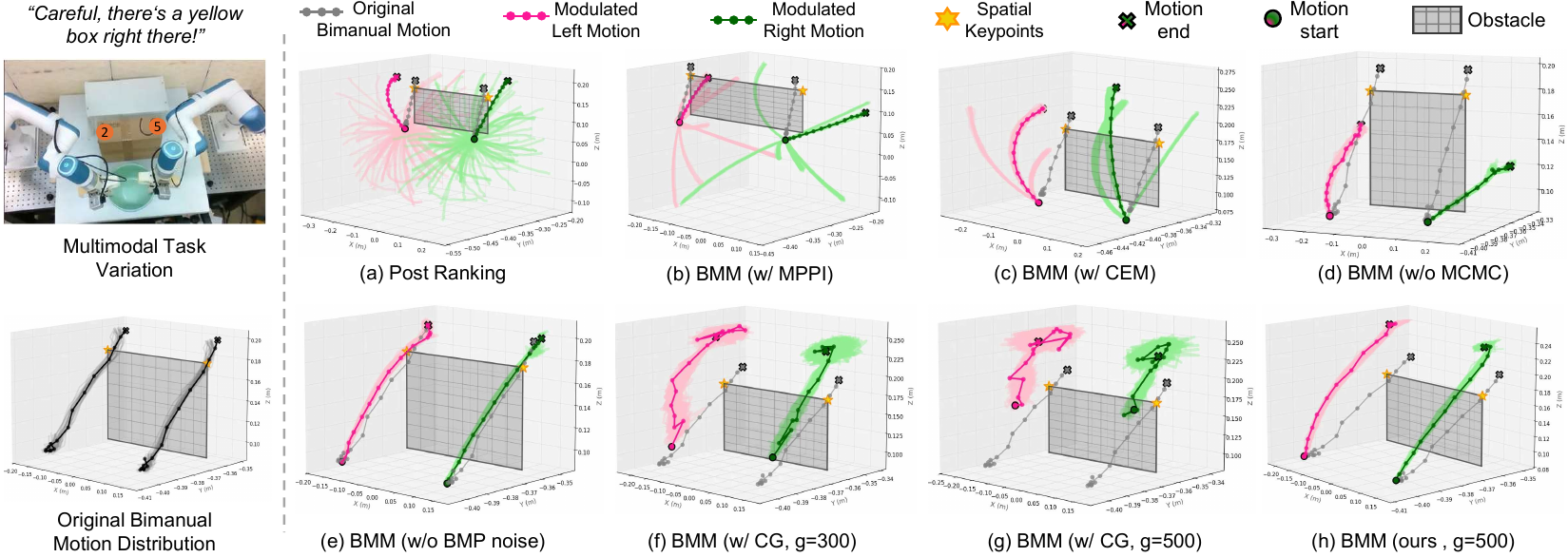}
\caption{\textbf{Qualitative Results of Online Bimanual Motion Modulation.} Left: Task variation and original bimanual motion distribution. Right: Comparison of modulated motions from various baselines. Our method (h) achieves the best balance of intent alignment, task compatibility, and dual-arm coordination.}
\label{fig10}
\end{figure*}

\begin{figure}[htbp]
\centering
\includegraphics[width=0.85\linewidth]{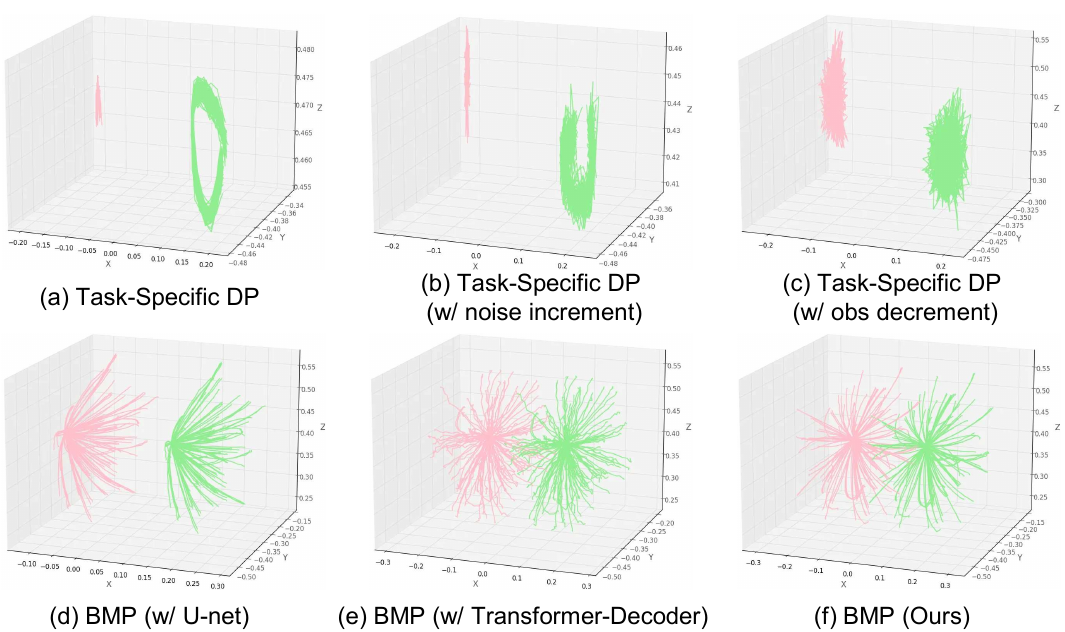}
\caption{\textbf{Qualitative Results of BMP Evaluation.} (a)–(c): Motions sampled from task-specific DP under different settings. (d)–(e): Motions sampled from BMP architectural variants. (f): Our BMP generates motion proposals that balance spatial diversity, temporal smoothness, and sampling efficiency.}
\label{BMP qualitative results}
\end{figure}

As shown in Table~\ref{ESA-COT quantitatives}, ESA-CoT consistently outperforms all baselines and ablations. Few-shot Prompting yields the lowest RA and ER, underscoring the importance of multimodal reasoning for interpreting implicit adaptation objectives. While Vanilla CoT improves over Few-shot Prompting, it remains markedly inferior to ESA-CoT, indicating that embodiment augmentation enhances both the reasoning accuracy and physical feasibility. Full ESA-CoT achieves the highest performance among all variants, validating the necessity of each component in the reasoning chain. Notably, ESA-CoT (w/o BMT) shows a significant RA drop, confirming the benefit of incorporating bimanual domain knowledge. ESA-CoT (w/ Diff Reward) exhibits a clear decline in ER, mainly due to execution failures induced by the inferred differentiable reward, highlighting the advantage of using black-box adaptation objectives. Qualitative results in Figure~\ref{ESA-COT qualitatives} demonstrate that ESA-CoT infers feasible rewards and coordination across diverse task variations.

\subsection{Evaluation of Bimanual Motion Prior}\label{sec.V-C}
We first evaluate the necessity of introducing BMP to generate initial proposals, as opposed to sampling directly from task-specific motion distributions. We compare BMP sampling against: 1) DP sampling: samples initial motion proposals from the diffusion policy trained on the plate-wiping task; 2) DP sampling (w/ noise increment): increases noise level to promote diversity; 3) DP sampling (w/ obs decrement): weakens observation conditioning to encourage variability. As shown in Figure \ref{BMP qualitative results}, proposals sampled from the task-specific DP exhibit limited diversity and overfitting to demonstrations (i.e., wiping movement). Although increasing noise or reducing observation conditioning introduces variability (Figure~\ref{BMP qualitative results}(b)-(c)), it often compromises smoothness, leading to jerky or physically implausible motions. In contrast, BMP learns a task-agnostic bimanual motion prior that generates diverse and feasible proposals (Figure~\ref{BMP qualitative results}(f)), enabling more flexible and generalizable adaptation beyond demonstrations.

We then evaluate the design of BMP by comparing it against different architectural variants (see Table~\ref{tab2}). Three quantitative metrics are employed for evaluation: 1) FID~\cite{heusel2017gans}: measures the distributional distance between generated and ground-truth motions; 2) Diversity: captures spatial variability by quantifying how well the generated motions explore the dual-arm workspace; 3) Smoothness: measures the temporal continuity of motion sequences. U-Net baseline performs the worst, with the highest FID and lowest diversity. We attribute this to the limited receptive field of CNNs, which hinders modeling of long-term bimanual motion dependencies (Figure \ref{BMP qualitative results}(d)). Transformer decoder-only architecture improves diversity but exhibits poor smoothness, as the absence of an encoder limits temporal context aggregation, resulting in discontinuous motions (Figure \ref{BMP qualitative results}(e)). While the full Transformer encoder-decoder achieves high diversity and low FID, it requires more than twice the inference time compared to ours. Overall, our designed BMP achieves a favorable balance between diversity, smoothness, and efficiency.

\begin{figure}[!t]
\centering
\includegraphics[width=0.85\linewidth]{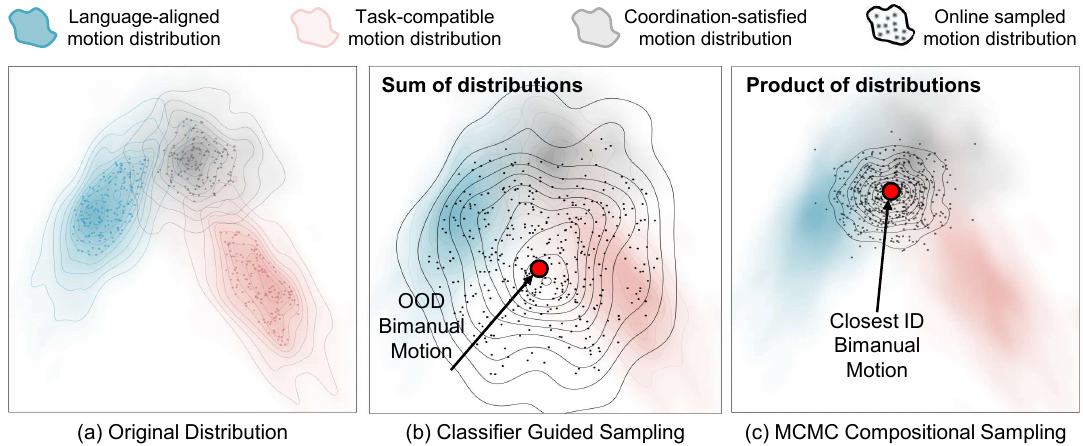}
\caption{\textbf{Comparison of MCMC and CG sampling.} Bimanual motions are projected to 2D using PCA and visualized via KDE \cite{waskom2021seaborn}. (a) Original component distributions. (b) CG sampling produces OOD motions. (c) Our MCMC sampling yields motions near the intersection of all components, avoiding distribution shift.}
\label{CG vs MCMC qualitatives}
\end{figure}

\begin{figure*}[!t]
\centering
\includegraphics[width=0.92\linewidth]{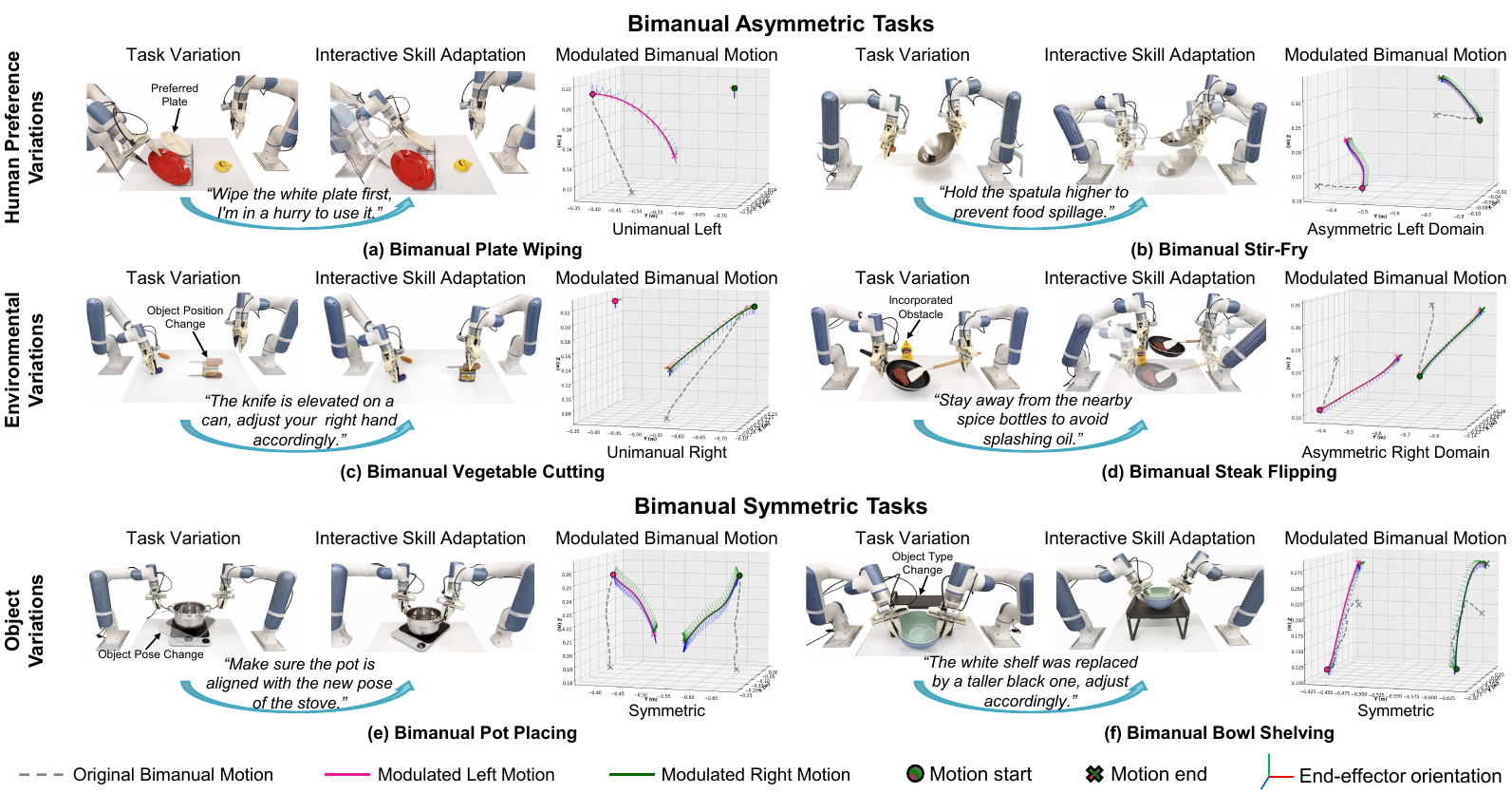}
\caption{\textbf{Real Robot Experiment Results.} BiSAIL demonstrates flexible and generalizable online interactive bimanual skill adaptation across diverse task variations, manipulation scenarios and coordination categories.}
\label{fig13}
\end{figure*}

\begin{figure}[!t]
\centering
\includegraphics[width=0.7\linewidth]{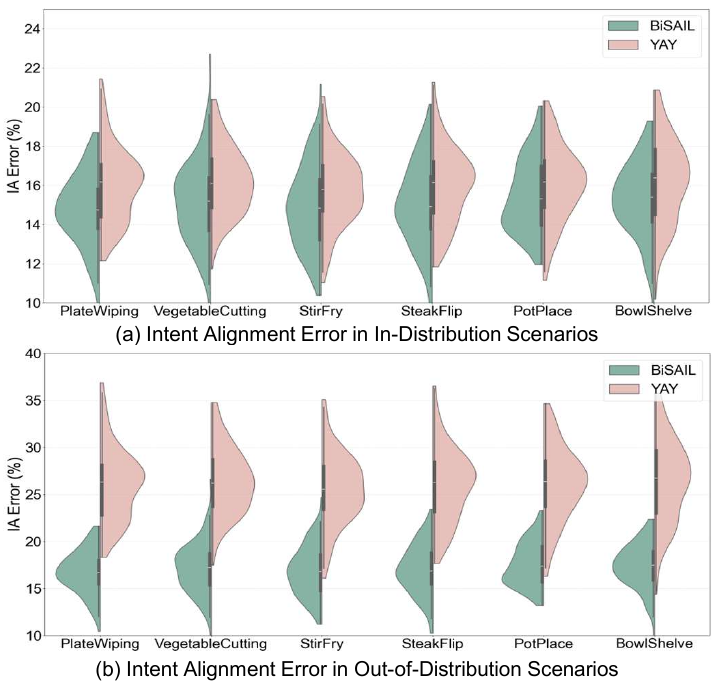}
\caption{\textbf{Comparison with End-to-End Adaptation Baseline.} (a) In ID settings, BiSAIL and YAY show comparable IA error. (b) In OOD settings, YAY shows a marked increase in IA error, while BiSAIL maintains low error.}
\label{fig:real_robot_experiments_error}
\vspace{-5pt}
\end{figure}

\begin{table}[!t]
\caption{Quantitative results for online bimanual motion modulation}
\label{tab:bmm_results}
\scriptsize
\centering
\begin{tabular*}{.5\textwidth}{@{\extracolsep{\fill}} lccc}
\toprule
Method & IA (\%) $\uparrow$ & TA (\%) $\uparrow$ & CS (\%) $\uparrow$ \\
\midrule
Post-hoc Ranking & 53.2 \scriptsize$\pm$ 1.1 & 40.5 \scriptsize$\pm$ 1.4 & 38.9 \scriptsize$\pm$ 2.2 \\
BMM (w/ MPPI) & 65.4 \scriptsize$\pm$ 1.7 & 68.2 \scriptsize$\pm$ 1.5 & 63.6 \scriptsize$\pm$ 2.8 \\
BMM (w/ CEM) & 71.6 \scriptsize$\pm$ 1.3 & 72.4 \scriptsize$\pm$ 2.1 & 70.2 \scriptsize$\pm$ 1.2 \\
BMM (w/o MCMC) & 76.5 \scriptsize$\pm$ 2.4 & 55.7 \scriptsize$\pm$ 2.6 & 58.3 \scriptsize$\pm$ 2.2 \\
BMM (w/ CG, g=300) & 78.9 \scriptsize$\pm$ 1.5 & 75.8 \scriptsize$\pm$ 1.3 & 72.1 \scriptsize$\pm$ 1.4 \\
BMM (w/ CG, g=500) & 64.7 \scriptsize$\pm$ 1.6 & 62.3 \scriptsize$\pm$ 2.7 & 64.5 \scriptsize$\pm$ 1.2 \\
BMM (w/o BMP noise) & 32.3 \scriptsize$\pm$ 2.2 & 61.8 \scriptsize$\pm$ 1.1 & 82.7 \scriptsize$\pm$ 1.3 \\
\midrule
BMM (ours) & \textbf{86.8 \scriptsize$\pm$ 1.2} & \textbf{82.5 \scriptsize$\pm$ 1.7} & \textbf{83.9 \scriptsize$\pm$ 1.6} \\
\bottomrule
\end{tabular*}
\end{table}

\subsection{Evaluation of Online Bimanual Motion Modulation}
We evaluate the proposed Bimanual Motion Modulation (BMM) algorithm against several ablated baselines: (i) Post-hoc Ranking selects the highest-scoring BMP proposals directly; (ii) BMM (w/ CEM) replaces iterative diffusion optimization with CEM-based search~\cite{rubinstein1997optimization}; (iii) BMM (w/ MPPI) replaces diffusion optimization with MPPI-based search~\cite{williams2015model}; (iv) BMM (w/o MCMC) removes the MCMC sampling process; (v) BMM (w/ CG) replaces MCMC with Classifier-Guided sampling~\cite{dhariwal2021diffusion}; (vi) BMM (w/o BMP noise) disables the injection of BMP noise during MCMC sampling to examine its effect on motion diversity. We conduct evaluations on the bowl shelving task using 10 adaptation objectives derived from ESA-CoT. Each method is tested in 20 independent trials. Three quantitative metrics adapted from~\cite{wang2025inference} are used: 1) Intent Alignment (IA) measures the percentage of modulated motions consistent with human feedback;
2) Task Alignment (TA) captures the percentage of motions that complete the intended task;
3) Constraint Satisfaction (CS) quantifies the percentage of motions satisfying coordination constraints.

As shown in Table~\ref{tab:bmm_results}, Post-hoc Ranking yields the lowest TA and CS, indicating the limitations of relying solely on initial BMP proposals without further modulation. BMM with CEM and MPPI report significantly higher IA error than ours, due to the degradation of Gaussian mutations in high-dimensional spaces. In contrast, our method enhances mutation efficiency via a truncated diffusion-denoising process that effectively explores high-dimensional multimodal bimanual motion priors. Removing MCMC sampling causes substantial drops in TA and CS, as modulated motions often fail to satisfy task requirements or dual-arm coordination (Figure~\ref{fig10}(d)). Although BMM (w/ CG) achieves reasonable CS, it suffers from distribution shift as the guidance ratio $g$ increases (Figure~\ref{fig10}(f)-(g)). In contrast, BMM (ours) consistently generates in-distribution motions under the same $g$. This is because CG approximates a sum of distributions, whereas MCMC targets their product (see Figure~\ref{CG vs MCMC qualitatives}). As a result, increasing $g$ causes CG to drift out of distribution, while MCMC identifies bimanual motions closest to the composed distribution. Finally, ablating BMP noise during compositional sampling causes the modulation to collapse back to task-specific behavior (Figure~\ref{fig10}(e)), confirming its role in preserving diversity. These results validate both the effectiveness of BMM and the necessity of its key components.

\begin{table*}[t]
\centering
\caption{Real-robot experiment results of online interactive bimanual skill adaptation}
\scriptsize
\resizebox{\textwidth}{!}{%
\begin{tabular}{lccccccc}
\toprule
\multicolumn{8}{c}{Object Variations} \\
\midrule
Method \textbackslash{} Tasks & PlateWiping & VegetableCutting & StirFry & SteakFlip & PotPlace & BowlShelve & Avg.TSR \\
\midrule
DP~\cite{chi2023diffusion}    & 0.33 \scriptsize$\pm$ 0.11 & 0.13 \scriptsize$\pm$ 0.09 & 0.27 \scriptsize$\pm$ 0.08 & 0.20 \scriptsize$\pm$ 0.09 & 0.33 \scriptsize$\pm$ 0.12 & 0.20 \scriptsize$\pm$ 0.11 & 0.24 \scriptsize$\pm$ 0.08 \\
ACT~\cite{zhao2023learning}  & 0.20 \scriptsize$\pm$ 0.07 & 0.20 \scriptsize$\pm$ 0.06 & 0.27 \scriptsize$\pm$ 0.09 & 0.13 \scriptsize$\pm$ 0.10 & 0.27 \scriptsize$\pm$ 0.10 & 0.33 \scriptsize$\pm$ 0.12 & 0.23 \scriptsize$\pm$ 0.07 \\
IDMP~\cite{wang2024cooperative} & 0.40 \scriptsize$\pm$ 0.11 & 0.33 \scriptsize$\pm$ 0.08 & 0.27 \scriptsize$\pm$ 0.07 & 0.40 \scriptsize$\pm$ 0.10 & 0.73 \scriptsize$\pm$ 0.11 & 0.67 \scriptsize$\pm$ 0.12 & 0.47 \scriptsize$\pm$ 0.18 \\
DSRL~\cite{wagenmaker2025steering} & 0.73 \scriptsize$\pm$ 0.04 & 0.67 \scriptsize$\pm$ 0.12 & \textbf{0.80 \scriptsize$\pm$ 0.09} & 0.67 \scriptsize$\pm$ 0.08 & \textbf{0.87 \scriptsize$\pm$ 0.09} & 0.80 \scriptsize$\pm$ 0.12 & 0.76 \scriptsize$\pm$ 0.08 \\
YAY~\cite{shi2024yell}  & 0.60 \scriptsize$\pm$ 0.13 & 0.47 \scriptsize$\pm$ 0.11 & 0.60 \scriptsize$\pm$ 0.13 & 0.53 \scriptsize$\pm$ 0.11 & 0.73 \scriptsize$\pm$ 0.13 & 0.73 \scriptsize$\pm$ 0.10 & 0.61 \scriptsize$\pm$ 0.05 \\
LATTE~\cite{bucker2022latte} & 0.53 \scriptsize$\pm$ 0.09 & 0.47 \scriptsize$\pm$ 0.07 & 0.60 \scriptsize$\pm$ 0.12 & 0.47 \scriptsize$\pm$ 0.11 & 0.53 \scriptsize$\pm$ 0.10 & 0.60 \scriptsize$\pm$ 0.08 & 0.53 \scriptsize$\pm$ 0.06 \\
BiSAIL (ours)     & \textbf{0.73 \scriptsize$\pm$ 0.09} & \textbf{0.67 \scriptsize$\pm$ 0.08} & 0.80 \scriptsize$\pm$ 0.12 & \textbf{0.73 \scriptsize$\pm$ 0.09} & 0.80 \scriptsize$\pm$ 0.12 & \textbf{0.87 \scriptsize$\pm$ 0.09} & \textbf{0.77 \scriptsize$\pm$ 0.07} \\
\midrule
\multicolumn{8}{c}{Environmental Variations} \\
\midrule
Method \textbackslash{} Tasks & PlateWiping & VegetableCutting & StirFry & SteakFlip & PotPlace & BowlShelve & Avg.TSR \\
\midrule
DP~\cite{chi2023diffusion}    & 0.33 \scriptsize$\pm$ 0.09 & 0.13 \scriptsize$\pm$ 0.08 & 0.13 \scriptsize$\pm$ 0.09 & 0.13 \scriptsize$\pm$ 0.03 & 0.20 \scriptsize$\pm$ 0.13 & 0.20 \scriptsize$\pm$ 0.05 & 0.19 \scriptsize$\pm$ 0.06 \\
ACT~\cite{zhao2023learning}  & 0.20 \scriptsize$\pm$ 0.12 & 0.07 \scriptsize$\pm$ 0.09 & 0.13 \scriptsize$\pm$ 0.06 & 0.20 \scriptsize$\pm$ 0.09 & 0.13 \scriptsize$\pm$ 0.09 & 0.20 \scriptsize$\pm$ 0.11 & 0.16 \scriptsize$\pm$ 0.05 \\
IDMP~\cite{wang2024cooperative} & 0.33 \scriptsize$\pm$ 0.12 & 0.27 \scriptsize$\pm$ 0.11 & 0.40 \scriptsize$\pm$ 0.11 & 0.33 \scriptsize$\pm$ 0.12 & 0.67 \scriptsize$\pm$ 0.08 & 0.60 \scriptsize$\pm$ 0.09 & 0.43 \scriptsize$\pm$ 0.16 \\
DSRL~\cite{wagenmaker2025steering} & 0.73 \scriptsize$\pm$ 0.09 & 0.67 \scriptsize$\pm$ 0.12 & \textbf{0.71 \scriptsize$\pm$ 0.09} & 0.67 \scriptsize$\pm$ 0.10 & 0.80 \scriptsize$\pm$ 0.12 & 0.73 \scriptsize$\pm$ 0.06 & 0.72 \scriptsize$\pm$ 0.08 \\
YAY~\cite{shi2024yell}  & 0.60 \scriptsize$\pm$ 0.12 & 0.53 \scriptsize$\pm$ 0.10 & 0.73 \scriptsize$\pm$ 0.08 & 0.60 \scriptsize$\pm$ 0.11 & 0.53 \scriptsize$\pm$ 0.12 & 0.73 \scriptsize$\pm$ 0.07 & 0.62 \scriptsize$\pm$ 0.05 \\
LATTE~\cite{bucker2022latte} & 0.53 \scriptsize$\pm$ 0.12 & 0.47 \scriptsize$\pm$ 0.11 & 0.53 \scriptsize$\pm$ 0.10 & 0.60 \scriptsize$\pm$ 0.11 & 0.60 \scriptsize$\pm$ 0.08 & 0.47 \scriptsize$\pm$ 0.07 & 0.53 \scriptsize$\pm$ 0.06 \\
BiSAIL (ours)     & \textbf{0.80 \scriptsize$\pm$ 0.05} & \textbf{0.73 \scriptsize$\pm$ 0.09} & 0.67 \scriptsize$\pm$ 0.10 & \textbf{0.73 \scriptsize$\pm$ 0.09} & \textbf{0.87 \scriptsize$\pm$ 0.09} & \textbf{0.80 \scriptsize$\pm$ 0.12} & \textbf{0.77 \scriptsize$\pm$ 0.07} \\
\midrule
\multicolumn{8}{c}{Human Preference Variations} \\
\midrule
Method \textbackslash{} Tasks & PlateWiping & VegetableCutting & StirFry & SteakFlip & PotPlace & BowlShelve & Avg.TSR \\
\midrule
DP~\cite{chi2023diffusion}    & 0.27 \scriptsize$\pm$ 0.06 & 0.20 \scriptsize$\pm$ 0.13 & 0.33 \scriptsize$\pm$ 0.08 & 0.20 \scriptsize$\pm$ 0.07 & 0.27 \scriptsize$\pm$ 0.11 & 0.40 \scriptsize$\pm$ 0.13 & 0.28 \scriptsize$\pm$ 0.07 \\
ACT~\cite{zhao2023learning}  & 0.27 \scriptsize$\pm$ 0.12 & 0.20 \scriptsize$\pm$ 0.10 & 0.27 \scriptsize$\pm$ 0.08 & 0.33 \scriptsize$\pm$ 0.09 & 0.33 \scriptsize$\pm$ 0.07 & 0.20 \scriptsize$\pm$ 0.12 & 0.27 \scriptsize$\pm$ 0.06 \\
IDMP~\cite{wang2024cooperative} & 0.27 \scriptsize$\pm$ 0.08 & 0.33 \scriptsize$\pm$ 0.10 & 0.40 \scriptsize$\pm$ 0.11 & 0.40 \scriptsize$\pm$ 0.07 & 0.60 \scriptsize$\pm$ 0.12 & 0.73 \scriptsize$\pm$ 0.09 & 0.46 \scriptsize$\pm$ 0.18 \\
DSRL~\cite{wagenmaker2025steering} & \textbf{0.87 \scriptsize$\pm$ 0.12} & 0.67 \scriptsize$\pm$ 0.09 & 0.67 \scriptsize$\pm$ 0.13 & \textbf{0.80 \scriptsize$\pm$ 0.07} & 0.73 \scriptsize$\pm$ 0.09 & 0.80 \scriptsize$\pm$ 0.09 & 0.76 \scriptsize$\pm$ 0.03 \\
YAY~\cite{shi2024yell}  & 0.73 \scriptsize$\pm$ 0.12 & 0.47 \scriptsize$\pm$ 0.11 & 0.53 \scriptsize$\pm$ 0.13 & 0.60 \scriptsize$\pm$ 0.13 & 0.47 \scriptsize$\pm$ 0.12 & 0.73 \scriptsize$\pm$ 0.09 & 0.59 \scriptsize$\pm$ 0.06 \\
LATTE~\cite{bucker2022latte} & 0.60 \scriptsize$\pm$ 0.11 & 0.47 \scriptsize$\pm$ 0.12 & 0.47 \scriptsize$\pm$ 0.12 & 0.53 \scriptsize$\pm$ 0.09 & 0.60 \scriptsize$\pm$ 0.08 & 0.47 \scriptsize$\pm$ 0.04 & 0.52 \scriptsize$\pm$ 0.04 \\
BiSAIL (ours)     & 0.73 \scriptsize$\pm$ 0.09 & \textbf{0.67 \scriptsize$\pm$ 0.07} & \textbf{0.80 \scriptsize$\pm$ 0.06} & 0.73 \scriptsize$\pm$ 0.09 & \textbf{0.80 \scriptsize$\pm$ 0.04} & \textbf{0.87 \scriptsize$\pm$ 0.09} & \textbf{0.76 \scriptsize$\pm$ 0.02} \\
\bottomrule
\end{tabular}
}
\label{tab:real_robot_experiments}
\end{table*}

\subsection{Real-Robot Experiments}
To evaluate the overall performance of BiSAIL, we conduct real-robot experiments across six bimanual tasks, comparing it against baselines from four categories. (1) Non-adaptive IL baselines, including DP\cite{chi2023diffusion} and ACT~\cite{zhao2023learning}. (2) A classical trajectory modulation baseline, IDMP~\cite{wang2024cooperative}, which extends the DMP framework to dual-arm settings via a hand-crafted coupling term. To handle multimodal task variations, it is augmented with ESA-CoT for inferring spatial keypoints as explicit adaptation objectives. (3) A policy fine-tuning baseline, DSRL~\cite{wagenmaker2025steering}, which employs reinforcement learning to fine-tune a diffusion policy for task adaptation. (4) End-to-end language-guided adaptation baselines, including YAY~\cite{shi2024yell} and LATTE \cite{bucker2022latte}, which are trained on large-scale datasets for language-conditioned skill modulation. All methods are evaluated under three common types of task variations in human-centric environments~\cite{cui2021toward}: Object Variations, Environmental Variations, and Human Preference Variations. For each variation type, we compose 15 random language feedback to guide the adaptation process.

\begin{table}[htbp]
\caption{Quantitative results of cross-embodiment evaluation}
\label{tab:cross_embodiment_results}
\scriptsize
\centering
\begin{tabular*}{.45\textwidth}{@{\extracolsep{\fill}} lccc}
\toprule
Task & IA (\%) $\uparrow$ & TA (\%) $\uparrow$ & CS (\%) $\uparrow$ \\
\midrule
SteakFlip & 78.3 \scriptsize$\pm$ 1.8 & 76.5 \scriptsize$\pm$ 2.2 & 75.2 \scriptsize$\pm$ 2.5 \\
VegetableCutting & 71.5 \scriptsize$\pm$ 1.4 & 73.3 \scriptsize$\pm$ 1.9 & 67.8 \scriptsize$\pm$ 1.1 \\
PlateWiping & 82.7 \scriptsize$\pm$ 2.2 & 84.2 \scriptsize$\pm$ 1.9 & 83.6 \scriptsize$\pm$ 2.3 \\
StirFry & 80.4 \scriptsize$\pm$ 2.6 & 78.9 \scriptsize$\pm$ 1.0 & 79.3 \scriptsize$\pm$ 1.4 \\
\bottomrule
\end{tabular*}
\end{table}

The average Task Success Rate (TSR) across all tasks and variation types is shown in Table~\ref{tab:real_robot_experiments}. Non-adaptive IL baselines yield the lowest TSR, confirming their inability to generalize beyond demonstrated settings. IDMP performs reasonably on symmetric tasks but fails in asymmetric ones due to its hand-crafted coupling term, reflecting the limited flexibility of traditional trajectory modulation under complex coordination changes. In contrast, BiSAIL achieves consistently high TSR across both task types by leveraging compositional motion sampling to flexibly incorporate diverse coordination constraints. Notably, BiSAIL matches the performance of DSRL while requiring no additional fine-tuning data, demonstrating strong zero-shot adaptation. End-to-end baselines perform well in-distribution but degrade significantly on OOD variations, as shown in Figure~\ref{fig:real_robot_experiments_error}. BiSAIL generalizes more effectively to unseen scenarios, enabled by its decoupled hierarchical architecture for interpreting novel task variations. Qualitatively, BiSAIL demonstrates flexible and generalizable online interactive bimanual skill adaptation across diverse task variations and coordination categories (Figure~\ref{fig13}). 

\begin{figure}[!t]
\centering
\includegraphics[width=0.95\linewidth]{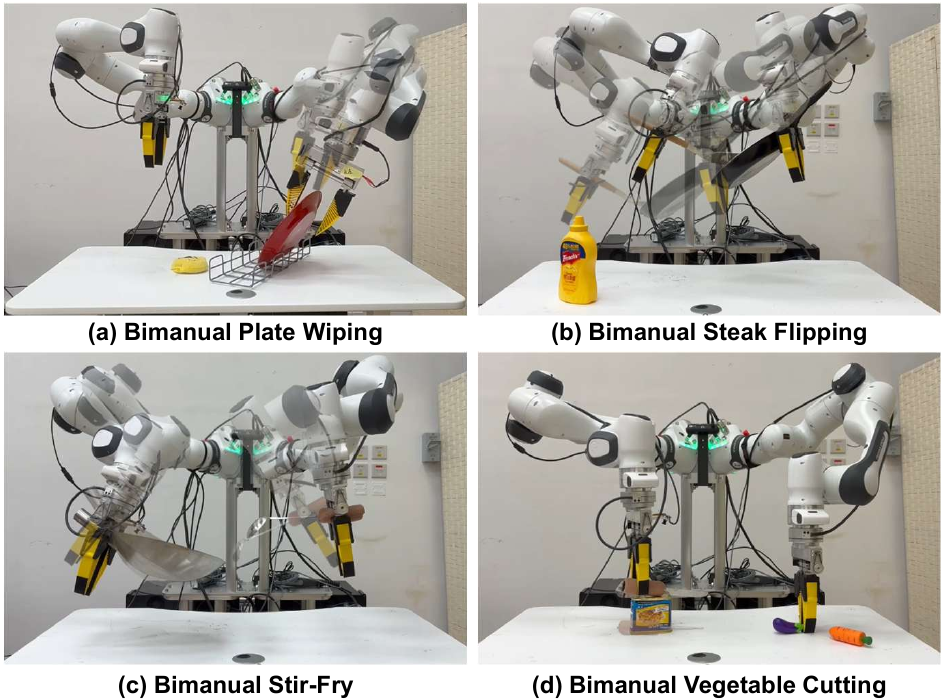}
\caption{\textbf{Cross-Embodiment Experiment Results.} BiSAIL consistently adapts to four bimanual tasks on a dual-arm Franka platform.}
\label{fig:franka_cross_embodiment}
\end{figure}
\vspace{-5pt}

\subsection{Cross-Embodiment Experiments}
To assess the scalability of BiSAIL across different robot embodiments, we conduct additional experiments on a dual-arm Franka platform, performing four asymmetric bimanual tasks (see Figure~\ref{fig:franka_cross_embodiment}). The evaluation protocol for each task follows the same setup as in Section~V-E. We retrain BMP using data generated specifically for the Franka robot, while keeping all other components unchanged. As shown in Table~V, BiSAIL consistently achieves high IA, TA, and CS scores across all tasks, demonstrating its robust cross-embodiment generalization. These results highlight the potential of BiSAIL as a plug-and-play bimanual skill adaptation framework for diverse dual-arm robotic platforms.

\section{Conclusion}
We presented BiSAIL, a novel online interactive adaptation framework that enables zero-shot generalization of learned bimanual skills to unseen deployment-time task variations. BiSAIL adopts a reason-then-modulate hierarchical paradigm that first reason adaptation objective from multimodal task variations and then coordinately adapts high-dimensional bimanual motions with online diffusion modulation. Through extensive real-world experiments, BiSAIL demonstrates strong generalization and flexibility across diverse task variations, coordination patterns, and robot embodiments. A current limitation of BiSAIL is its focus on kinematic-level adaptation, which restricts adjustments to spatial motions. Expanding the framework to incorporate dynamic-level adaptation, including modulation of execution speed, acceleration, and interaction forces, is a promising direction for future work. 

\bibliographystyle{IEEEtran}
\bibliography{ref}

\end{document}